\documentclass[conference]{IEEEtran}
\IEEEoverridecommandlockouts
\usepackage{cite}
\usepackage{amsmath,amssymb,amsfonts}
\usepackage{algorithmic}
\usepackage{graphicx}
\usepackage{textcomp}
\usepackage{xcolor}
\def\BibTeX{{\rm B\kern-.05em{\sc i\kern-.025em b}\kern-.08em
    T\kern-.1667em\lower.7ex\hbox{E}\kern-.125emX}}

\begin{document}

\title{Framework and Model Analysis on Bengali Document Layout Analysis Dataset: BaDLAD\\
\thanks{DL Sprint 2.0\cite{b2} has provided this opportunity to extend our knowledge and research on bengali document analysis}
}

\author{\IEEEauthorblockN{Kazi Reyazul Hasan\IEEEauthorrefmark{1},
 Mubasshira Musarrat\IEEEauthorrefmark{2},
 Sadif Ahmed\IEEEauthorrefmark{3}, and
 Shahriar Raj\IEEEauthorrefmark{4}}
\IEEEauthorblockA{\textit{Computer Science Engineering} \\
\textit{Bangladesh University of Engineering and Technology}\\
Dhaka, Bangladesh} \\
}

\maketitle

\begin{abstract}
This study focuses on understanding Bengali Document Layouts using advanced computer programs: Detectron2, YOLOv8, and SAM. We looked at lots of different Bengali documents in our study. Detectron2 is great at finding and separating different parts of documents, like textboxes and paragraphs. YOLOv8 is good at figuring out different tables and pictures. We also tried SAM, which helps us understand tricky layouts. We tested these programs to see how well they work. By comparing their accuracy and speed, we learned which one is good for different types of documents. Our research helps make sense of complex layouts in Bengali documents and can be useful for other languages too.
\end{abstract}


\section{Introduction}
Understanding the layout of documents is vital for extracting organized information from a jumble of text and pictures. Modern technology, especially deep learning, has made great strides in this area. One standout tool is Detectron2, made by Facebook AI Research (FAIR). It's really good at separating paragraphs and textboxes, but it initially struggled with spotting tables and images. However, with more training, it got better at images too.

We took Detectron2 and tried to make it even better. We played around with different tricks like changing how we trained it, using different pictures to teach it, and even using special operations to adjust its "masks" (the parts it finds in the documents).

We also checked out another tool called YOLOv8, which was expected to work well based on the original BaDLAD\cite{b1} research paper. But when we compared it with Detectron2, YOLOv8 didn't do as well, which was a surprise. So, we came up with a new idea. We used YOLOv8 to draw the outlines, and then we used another tool called SAM to fill in the colors. This might sound like extra work, but SAM is kind of new, and we wanted to see if it could help make the outlines better.

Our main goal was simple: make Detectron2 work better with complicated things like tables and images, and see if YOLOv8 and SAM could team up to draw better outlines. Through our tests and experiments, we tried to figure out what works best for understanding Bengali documents and how these tools could be helpful for other language related projects too. 

Our experiments have context limitations, but they highlight possibilities. This research pushes document analysis, crossing language barriers and using various applications.

\section{Methodology}

Our task involved analyzing BaDLAD document images, segmenting layouts into paragraphs, textboxes, images, and tables using trained models. We encoded segmentation masks using Run-Length Encoding and organized them into a concise .csv format for submission. This process ensured accurate layout categorization and efficient data representation.

\subsection{Implementation of Detectron2}

To use the power of Detectron2 for Bengali Document Layout Analysis, a carefully tested methodology\cite{b3} was employed. The following steps outline the process and key decisions made:
\begin{itemize}

\item Data Conversion: The COCO annotation data, a widely used format for object detection and segmentation, was converted to a format that Detectron2 can readily understand. This ensured integration of the dataset into the Detectron2 framework.

\item Model Selection: To achieve superior accuracy, the model "R101-FPN" was chosen for image segmentation. This model is equipped with advanced instance segmentation capabilities, suited for complex layout analysis tasks. In the case of Detectron2 models, using a larger backbone like ResNet-101 generally leads to better performance compared to ResNet-50. The increased depth and complexity of the backbone can capture more intricate features and patterns, making the model more accurate in tasks like object detection and instance segmentation.

\begin{table}[htbp]
\caption{Instance Segmentation Model Comparsion for Detctron2}
\begin{center}
\begin{tabular}{|c|c|c|c|c|c|c|}
\hline
\textbf{\textit{Name}} & \textbf{\textit{lr sch}} & \textbf{\textit{infer. time}}& \textbf{\textit{train mem}}& \textbf{\textit{box AP}}& \textbf{\textit{mask AP}}\\
\hline
R50-FPN& 3x& 0.043& 3.4& 41.0& 37.2  \\
\hline
R101-FPN& 3x& 0.056& 4.6& 42.9& 38.6  \\
\hline
X101-FPN& 3x&  0.103& 7.2& 44.3& 39.5  \\
\hline
\end{tabular}
\label{tab1}
\end{center}
\end{table}

\textbf{Why R101-FPN?}
Better box AP and mask AP compared to R50-FPN, trading with bearable inference time increase. X101-FPN performs better in APs, however, the tradeoff in inference time is huge (almost twice) and there is a good chance of memory overflow. Hence, after considering these cases, we chose to proceed with R101-FPN.

\item Learning Rate (LR) and Hyperparameters: A learning rate (LR) of 0.008 was utilized for optimal model training. Hyperparameters like batch size were set to 10, and the iteration number was chosen as 11,000. Although higher iteration numbers, such as 20,000+, can yield more accurate results, the decision to avoid prolonged training time to keep a balance between accuracy and training time.

\item Data Augmentation: Various augmentation techniques were applied to the training dataset. Adjustments in brightness, saturation, contrast, hue, along with horizontal and vertical flipping, cropping, and scaling, were implemented. These techniques enriched the model's ability to predict in diverse environments and varying conditions.

\item Preprocessing Considerations: While data augmentation enhanced the model's adaptability, certain preprocessing techniques were deliberately avoided during training, specifically binarization. This decision was taken for maintaining the model's versatility in real-world scenarios where images might not be binarized as well as keeping the pre-training period short.

\begin{figure}[htbp]
\centerline{\includegraphics[scale=0.5]{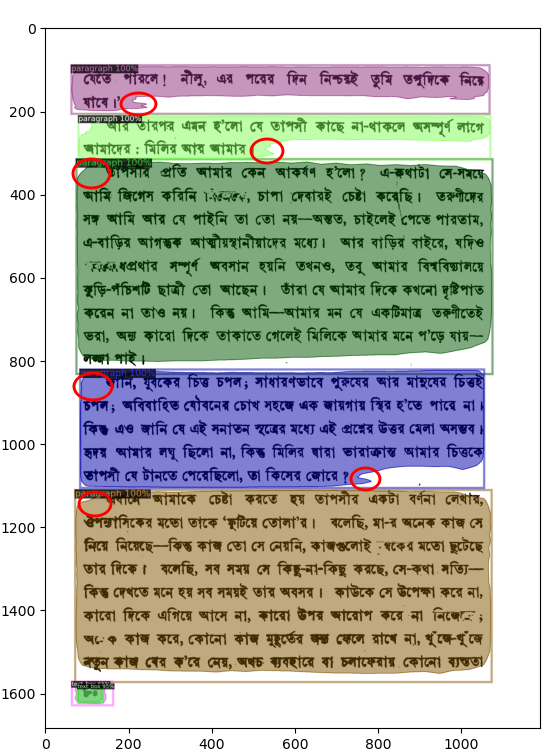}}
\caption{Mask detection example of Detectron2.}
\label{fig}
\end{figure}

\item Postprocessing Considerations: As the result masks deviated a bit from what was desired, a series of postprocessing techniques were employed to refine the results obtained from the Detectron2 framework. Erosion and dilation operations were applied to the segmented regions to fine-tune their boundaries. Erosion helped reduce noise and small artifacts, while dilation helped in expanding and connecting segmented areas for more cohesive results. A smoothing process was introduced to eliminate small fluctuations and irregularities in the segmented regions. However, smoothing made the masks way too heavy for rle encoding hence this idea was not continued later.
\end{itemize}

By carefully following this methodology, the framework was fine-tuned to address the complexities of Bengali Document Layout Analysis. The selected model, hyperparameters, and augmentation techniques collectively contributed to the model's ability to accurately segment and categorize different document components.

\subsection{Implementation of YOLOv8}
The YOLOv8 approach for Bengali Document Layout Analysis involved a specific set of techniques and parameters, tailored to optimize the performance of the YOLOv8 model. The following steps outline the methodology:
\begin{itemize}
 
\item Model Selection: The YOLOv8 model variant "yolov8m-seg.pt" was chosen as the basis for our analysis. This model's architecture and capabilities were well-suited for our document layout segmentation tasks.

\item Training Configuration:
\begin{enumerate} 
\item Epochs: The training process was conducted over 4 epochs. This limited number of epochs was selected to strike a balance between training time and achieving satisfactory results.
\item Basic Augmentation: Basic data augmentation techniques were applied during training. These techniques, such as random cropping, resizing, and flipping, help improve model generalization and robustness.
\item Cosine Learning Rate (Cos LR): The Cos LR scheduler was employed, which gradually reduces the learning rate as training progresses. This approach helps the model converge to optimal solutions by fine-tuning its parameters more effectively.
\item Label Smoothing: Label smoothing was implemented as part of the training process. Label smoothing mitigates the model's tendency to become overconfident in its predictions by slightly adjusting target labels. This helps in improving the model's generalization and reduces the risk of overfitting.
\end{enumerate}

\item Learning Rate and Optimization:
\begin{enumerate} 
\item Initial Learning Rate (lr0): An initial learning rate (lr0) of 0.01 was set to regulate the weight updates during the training process.
\item Learning Rate Factor (lrf): A learning rate factor (lrf) of 0.001 was used to adjust the learning rate during the cosine learning rate decay.
\end{enumerate}
\end{itemize}

\subsection{Implementation of SAM (Segment Anything)}

The SAM framework\cite{b5} usage for Bengali Document Layout Analysis employed a unique approach, combining pre-trained models and a new integration process. The following steps outline the methodology:
\begin{itemize} 
\item Pretrained SAM Model: We began by using a pre-trained SAM model on our Bengali Document Layout Analysis Dataset (BaDLAD). This pre-trained model offered a starting point for understanding the to what extent it can read Bengali documents.
\item YOLOv8 for Bounding Boxes: YOLOv8 was employed for image detection rather than segmentation. The model was trained to identify and locate various document components, providing bounding box coordinates.
\item Inference for Bounding Boxes: After training YOLOv8 for image detection, the model was used for inference on the BaDLAD dataset. This process resulted in accurate bounding box predictions around different elements within the documents.
\item Integration with SAM: An innovative approach was introduced by integrating YOLOv8's bounding box predictions with the capabilities of SAM. SAM was employed to generate masks over the bounding boxes, in hopes of segmenting the identified document components.
\end{itemize}

\section{Results and Findings}
Our thorough exploration using Detectron2, YOLOv8, and SAM in analyzing Bengali Document Layouts uncovered significant insights, giving us insights of their strengths and limitations. All the models were trained based on BaDLAD training dataset. The given instances of the selected 19346 images among 4 categories were:
\begin{table}[htbp]
\caption{Instances of Training Dataset}
\begin{center}
\begin{tabular}{|c|c|}
\hline
\textbf{\textit{Category}} & \textbf{\textit{Instances}} \\
\hline
paragraph& $198933$  \\
\hline
textbox& $194113$  \\
\hline
image& $9737$  \\
\hline
table& $1297$\\
\hline
total& $404080$  \\
\hline
\end{tabular}
\label{tab1}
\end{center}
\end{table}

\begin{itemize}
     
\item Detectron2's Better Performance: Detectron2, our leading model, showcased remarkable accuracy across diverse document elements like textboxes, paragraphs, images and tables.
While its mask drawing accuracy was slightly lower than YOLOv8, Detectron2's comprehensive layout analysis capabilities, especially following extensive training, established its supremacy.

\item Postprocessing Findings: The selection of morphological operations for use with Detectron2 did not yield significant improvements. While our research suggested that applying opening and closing\cite{b4} could potentially help reduce unnecessary mask extensions, the actual results showed a slight underperformance in terms of dice scores. Despite our initial expectations, the practical application of these operations did not lead to the desired enhancement in mask accuracy.

\begin{figure}[htbp]
\centerline{\includegraphics[scale=0.3]{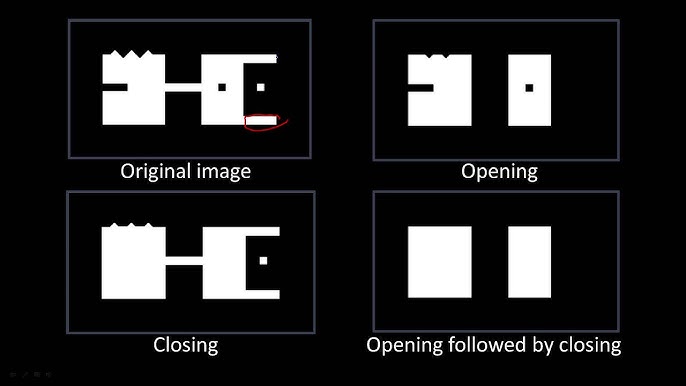}}
\caption{Opening and closing was applied sequentially on masks.}
\label{fig}
\end{figure}

\item YOLOv8's Abilities and Challenges: YOLOv8 exhibited superior accuracy in mask drawing, particularly excelling in specific components.
Surprisingly, YOLOv8 struggled to precisely identify crucial elements like textboxes, paragraphs, and images. This limitation impacted its overall performance despite its strong mask drawing competence.

\begin{figure}[htbp]
\centerline{\includegraphics[scale=0.55]{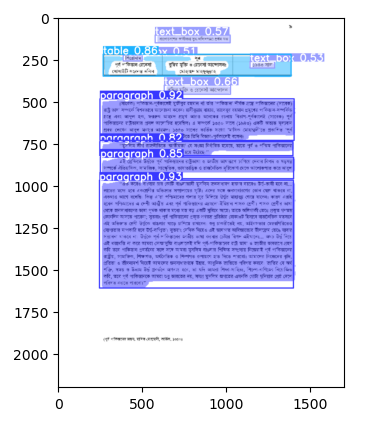}}
\caption{Mask detection example of YOLOv8.}
\label{fig}
\end{figure}

\begin{figure}[htbp]
\centerline{\includegraphics[scale=0.40]{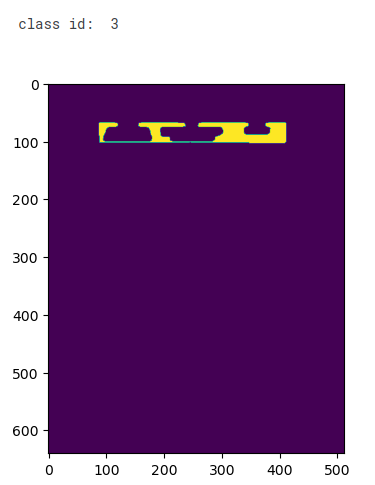}}
\caption{Masks remain unfilled for tables in YOLOv8.}
\label{fig}
\end{figure}

\item Handling Image: Working with images proved more complex compared to handling paragraphs. Many images suffered from issues like text overlapping or dark areas, causing them to break apart. 
\begin{figure}[htbp]
\centerline{\includegraphics[scale=0.5]{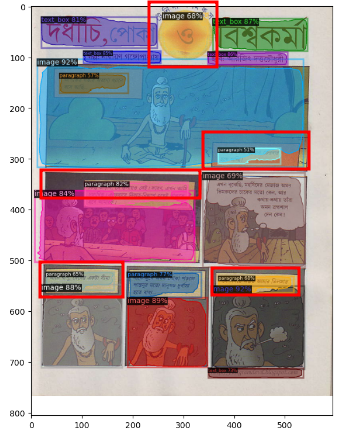}}
\caption{Image masks are irregular, high inaccuracy visible.}
\label{fig}
\end{figure}

To solve this, we came up with a clever idea: we used the cv2 library to create masks that follow the shapes of images, like polygons. This library is really good at working with images, so it might help fix the problem of image masks breaking apart. By using cv2 to make better image masks, we hope to improve how we understand images within a document in document analysis.

\item Integration Hurdles with SAM: Our innovative fusion of YOLOv8's bounding box predictions with SAM's segmentation yielded mixed outcomes.

SAM's lack of familiarity with Bengali documents played a role in its modest performance, underscoring the significance of tailored pre-training.

\item Extended Training Elevates Detectron2: Prolonged training significantly elevated Detectron2's performance, as evidenced by its ability to discern varied document layouts with enhanced accuracy.

\item Deviation from Prior Expectations: Our findings diverged from predictions laid out in the BaDLAD research paper, where YOLOv8 held a higher ranking. Unique dataset characteristics and layout complexities may have contributed to this variance.
\item 
Practical Implications and Future Directions: The implications of our discoveries extend beyond Bengali documents, offering insights applicable to diverse languages and sectors.
Future research could focus on enhancing SAM's efficacy by tailoring its training to Bengali documents, unlocking its untapped potential. Frameworks such as MMDetection are worthy of more exploring as well.
\end{itemize}

\begin{table}[htbp]
\caption{Dice Scores of different Frameworks (9 hours training approx.)}
\begin{center}
\begin{tabular}{|c|c|}
\hline
\textbf{\textit{Framework}} & \textbf{\textit{Score}} \\
\hline
Detectron2& $0.86328$  \\
\hline
Detectron2 (morphological operations)& $0.86284$  \\
\hline
YOLOv8& $0.49584$  \\
\hline
YOLOv8 with SAM& $0.46343$\\
\hline
\end{tabular}
\label{tab1}
\end{center}
\end{table}
In conclusion, our results shed light on the relation between accuracy and mask drawing abilities within Detectron2, YOLOv8, and SAM for Bengali Document Layout Analysis. Detectron2's comprehensive analysis, YOLOv8's nuanced capabilities, and SAM's integration challenges collectively enrich our comprehension of their functionalities and pave the way for potential enhancements.It is worthy of notation that the discovered results are highly dependent on training time period, quality of test images and the valued expertise of models.

\section{Discussion}
Let's talk about what we discovered while exploring Detectron2, YOLOv8, and SAM for understanding Bengali Document Layouts. Here's a simpler way to understand what we found:
\begin{itemize} 

\item Detectron2's Strength and Training Time: Detectron2 did a really good job at figuring out different parts of the documents. It got even better after training it for a long time. This was a bit surprising because it did better than what was expected based on previous research. However, training it for a long time takes more time.

\item YOLOv8's Special Skill and Challenges: YOLOv8 was very good at drawing masks, like coloring inside the lines. But it struggled to find important things like textboxes and paragraphs accurately. This means it's good at one thing but not so great at the overall job.

\item SAM's New Idea and Need for Training: SAM combined with YOLOv8 was a new and creative idea. However, it didn't work as well as we hoped. This might be because SAM wasn't familiar with Bengali documents. If we train SAM on Bengali documents, it might become more useful.

\begin{figure}[htbp]
\centerline{\includegraphics[scale=0.23]{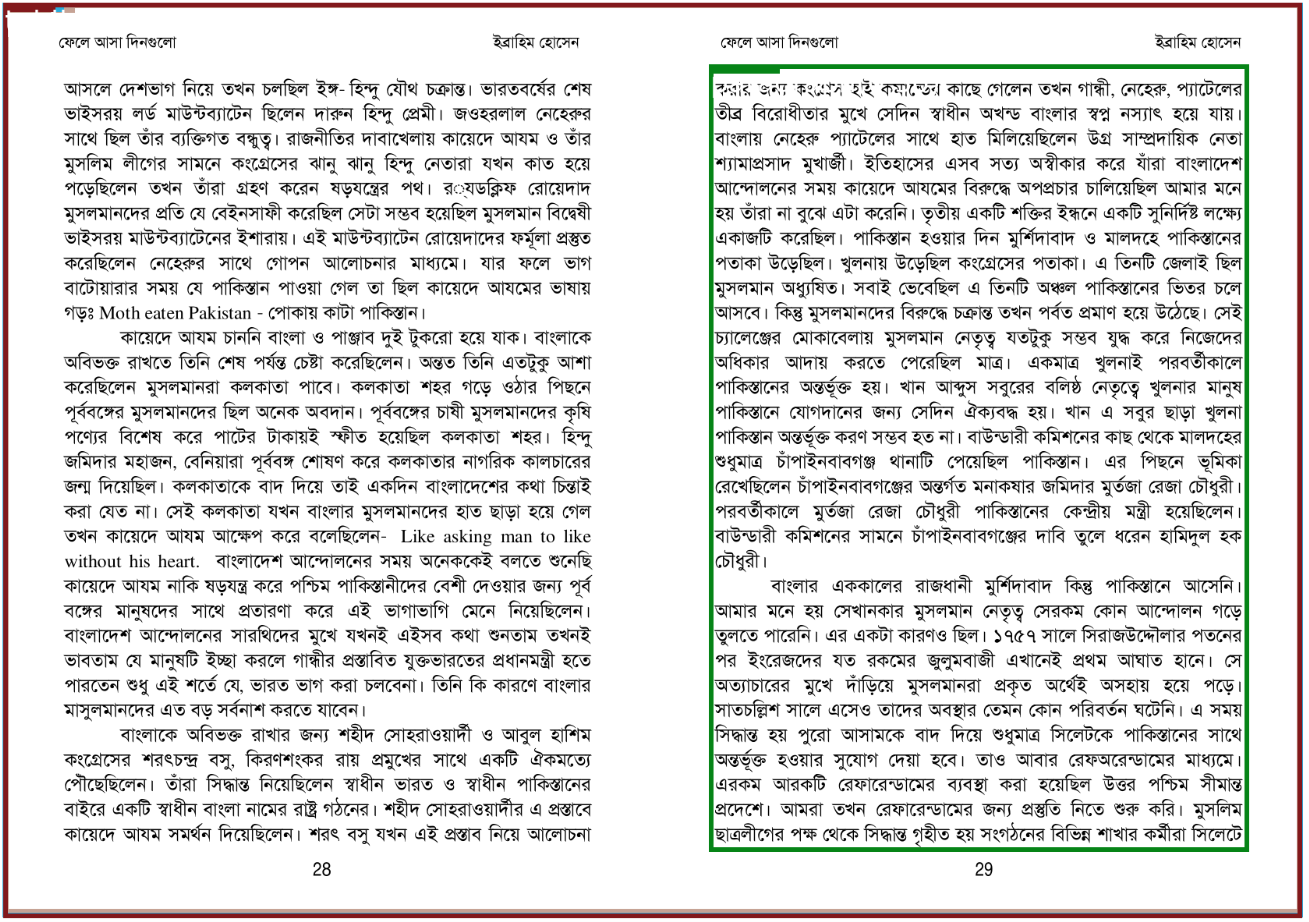}}
\caption{SAM failing to detect easiest of bangla paragraphs.}
\label{fig}
\end{figure}

\item Hyperparameters played a crucial role in model performance. A deeper exploration into fine-tuning hyperparameters could shed light on their impact on document layout analysis. This discussion could revolve around questions like: What learning rate works best for each model? How does adjusting batch size affect convergence and accuracy? Exploring the effects of various hyperparameters, such as optimizer choices, weight decay, and learning rate schedules, patterns like cosine learning rate, could uncover model performance specifically for document layout segmentation.

\item Another interesting aspect to consider is the potential of transfer learning and adaptation. Many pre-trained models are trained on generic datasets. Exploring how well these models can be fine-tuned or adapted to the specifics of the BaDLAD dataset, which contains Bengali documents, could reveal insights into the adaptability of AI models to different languages and content types.

\item The things we learned apply not only to Bengali documents but also to other languages and topics. Different methods have their strengths and weaknesses, so we can choose the right one for the job. SAM has potential if we train it better, and this research can help others who want to analyze documents.

\item What Comes Next: Our research gives practical advice for people who want to analyze document layouts. It also points to future possibilities, like improving SAM with better training to make it more useful for drawing masks accurately.

\end{itemize}

In the end, our discussion shows how Detectron2, YOLOv8, and SAM each have their own abilities and challenges. It's like having different tools in a toolbox, and depending on what we need, we can choose the best tool for the job. This research helps us understand these tools better and opens the door for more improvements in the future.
For our given scenario, we chose to stick with Detectron2 framework for best accuracy within time constraints.

\section{Conclusion}
Our journey through BaDLAD document analysis gave us important insights into different parts of the layout. Detectron2 stood out by accurately finding textboxes, paragraphs, images, and tables, especially after a lot of training. YOLOv8 was really good at drawing masks, but it sometimes struggled to recognize everything, showing it's good at one thing but not everything. Trying SAM with YOLOv8 was an interesting experiment, but it didn't work perfectly since SAM wasn't familiar with our kind of documents.

Our results surprised us because Detectron2 did even better than what our focused research paper said. Probably it is more applicable for mAP scores where Detectron2 may face issues. YOLOv8 has its own strong points, and SAM could be useful with more training. This isn't just about Bengali documents – we can use these ideas for other clasification of documents and topics too.

What we learned here helps people who want to understand document layouts better. This journey helps us know more about these methods and how they can help us figure out complex document structures. We are looking forward to explore model ensembling from here on, a new approach to find out a better conclusion. Model ensembling involves combining the predictions of multiple different models to improve overall performance and accuracy. In the context of document analysis, like our study, model ensembling could mean taking the outputs of Detectron2, YOLOv8, and SAM, and combining their results to make a more accurate final segmentation of the document layout. By combining the strengths of different models, we can overcome individual weaknesses and get a better overall result. This approach often leads to better accuracy and robustness in complex tasks like document layout analysis. However, the task is not easy and we could only figure out the strengths and weaknesses yet, the merging may take complicated steps which will require more time and research. 

\section*{Acknowledgement}

We express our sincere gratitude to Bangladesh University of Engineering and Technology (BUET) and Bengali.AI for their invaluable support. Their guidance, demo codes, and workshops have played an important role in enhancing our understanding of document analysis. BUET's academic environment and Bengali.AI's resources provided key insights, contributing significantly to our research. We acknowledge their contributions, which have greatly influenced our methodology and outcomes in this field.
We would also like to express our appreciation to the developers and contributors of Detectron2, YOLOv8, and Segment Anything (SAM), whose open-source documentation and GitHub repositories have been invaluable resources.

\vspace{12pt}

\end{document}